\documentclass[lettersize,journal]{IEEEtran}
\title{Bi-AQUA: Bilateral Control-Based Imitation Learning for Underwater Robot Arms via Lighting-Aware Action Chunking with Transformers}

\usepackage{pifont}
\usepackage{booktabs}
\usepackage{multirow}
\usepackage{tabularx}
\usepackage{epsfig} 
\usepackage{amssymb}
\usepackage{subcaption}
\usepackage{amsmath}
\usepackage{url}

\newcommand{\cmark}{\ding{51}}
\newcommand{\xmark}{\ding{55}}

\author{
Takeru Tsunoori$^{\dag 1}$, Masato Kobayashi$^{\dag 1,2*}$, Yuki Uranishi$^{1}$
\thanks{
$^{\dag}$ Equal Contribution, 
$^{1}$ The University of Osaka, 
$^{2}$ Kobe University,\\
$^*$ corresponding author: kobayashi.masato.cmc@osaka-u.ac.jp
}
}

\begin{document}

\maketitle
\thispagestyle{empty}
\pagestyle{empty}

\begin{abstract}
Underwater robotic manipulation remains challenging because lighting variation, color attenuation, scattering, and reduced visibility can severely degrade visuomotor policies. We present Bi-AQUA, the first underwater bilateral control-based imitation learning framework for robot arms that explicitly models lighting within the policy. Bi-AQUA integrates transformer-based bilateral action chunking with a hierarchical lighting-aware design composed of a label-free Lighting Encoder, FiLM-based visual feature modulation, and a lighting token for action conditioning. This design enables adaptation to static and dynamically changing underwater illumination while preserving the force-sensitive advantages of bilateral control, which are particularly important in long-horizon and contact-rich manipulation. Real-world experiments on underwater pick-and-place, drawer closing, and peg extraction tasks show that Bi-AQUA outperforms a bilateral baseline without lighting modeling and achieves robust performance under seen, unseen, and changing lighting conditions. These results highlight the importance of combining explicit lighting modeling with force-aware bilateral imitation learning for reliable underwater manipulation.
For additional material, please check: \url{https://mertcookimg.github.io/bi-aqua}
\end{abstract}


\section{Introduction}
Underwater robotic manipulation is highly challenging because scene appearance can change drastically within seconds due to shifts in underwater lighting spectrum, intensity, and direction~\cite{Yang2019AnIn-DepthSurveyofUnderwaterImageEnhancement}. Wavelength-dependent attenuation, scattering, turbidity, and bubbles disrupt visual consistency, often causing severe action drift in standard visuomotor policies. Although underwater image enhancement methods improve perceptual quality~\cite{Varghese2025Sea-InginLow-Light, Jamieson2023DeepSeeColor}, they do not address the core control problem: closed-loop policies must adapt both perception and action generation to rapidly changing illumination.
While traditional underwater manipulation relies on teleoperation, recent work explores RL and IL-based autonomy~\cite{lin2024UINAV, liu2024selfimprovingautonomousunderwatermanipulation, gu2025usimu0visionlanguageactiondataset}. However, existing approaches either depend purely on vision or employ unilateral control without force feedback, and none explicitly model lighting as a latent factor within the visuomotor policy.

\begin{figure}[t]
  \centering
   \includegraphics[width=\linewidth]{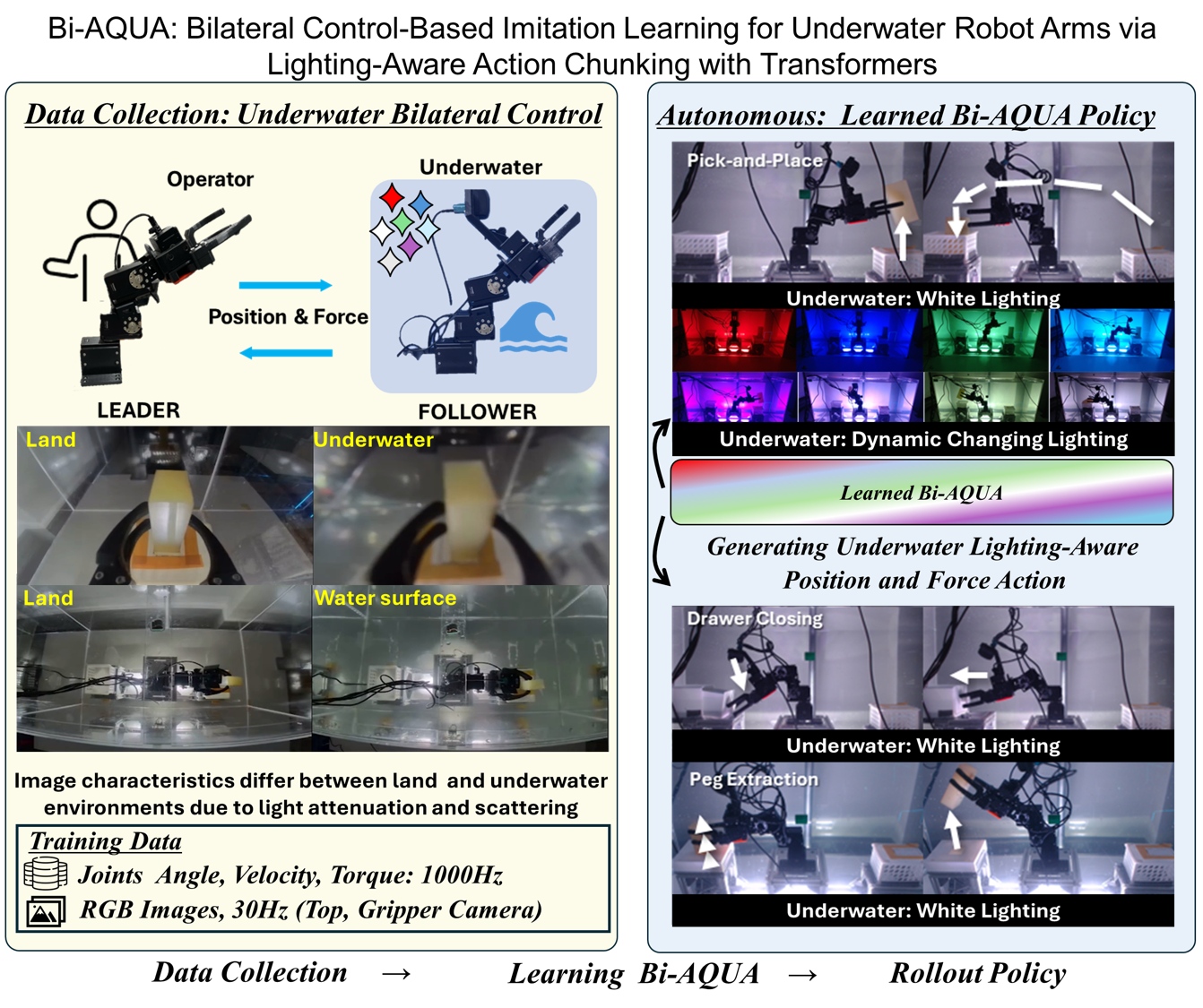}
   \caption{Concept of Bi-AQUA.}
   \label{fig:tsr}
\end{figure}

\begin{figure}[t]
  \centering
   \includegraphics[width=\linewidth]{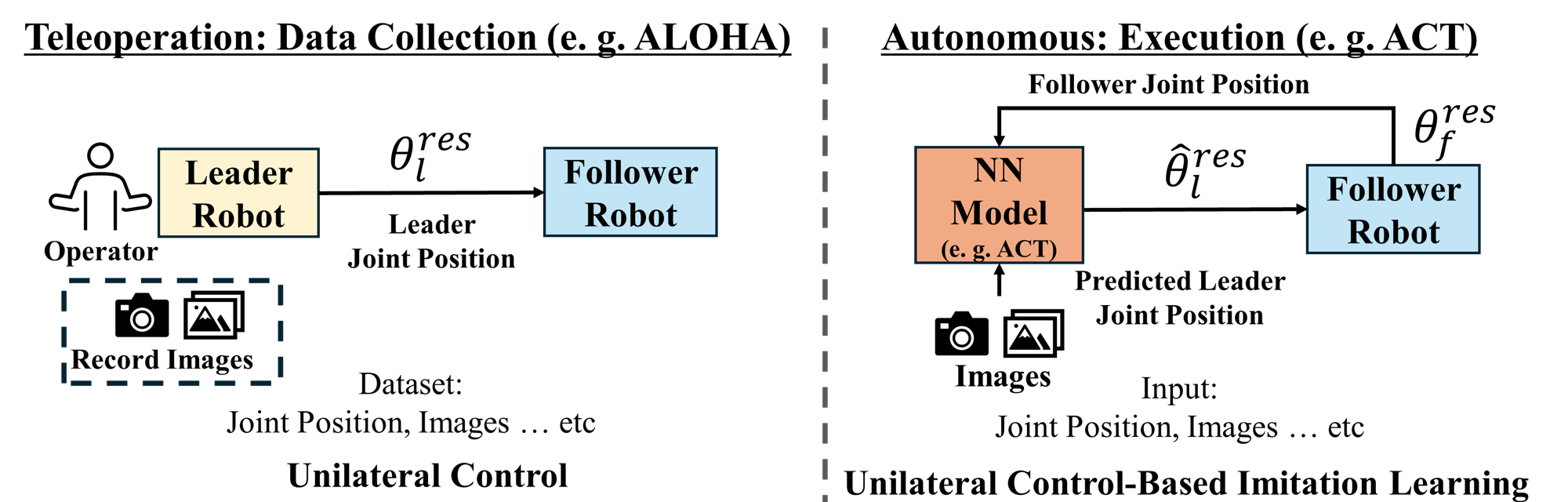}
   \caption{Unilateral Control-based Imitation Learning}
   \label{fig:uni}
\end{figure}

In terrestrial environments, imitation learning (IL) through leader--follower teleoperation has proven highly effective for acquiring complex visuomotor skills~\cite{tsuji2025surveyimitationlearningcontactrich}. 
Recent systems based on ALOHA and Action Chunking with Transformers (ACT)~\cite{zhao2023learningfinegrainedbimanualmanipulation, aloha2team2024aloha2enhancedlowcost}, 
Mobile ALOHA with ACT~\cite{fu2024mobilealohalearningbimanual}, 
and diffusion-based visuomotor policies~\cite{chi2024diffusionpolicyvisuomotorpolicy} achieve impressive real-world performance.
These method yet all rely on unilateral control, as shown in Fig.~\ref{fig:uni}. 
Without force feedback, these methods struggle in contact-rich or visually ambiguous interactions.

\begin{figure}[t]
  \centering
   \includegraphics[width=\linewidth]{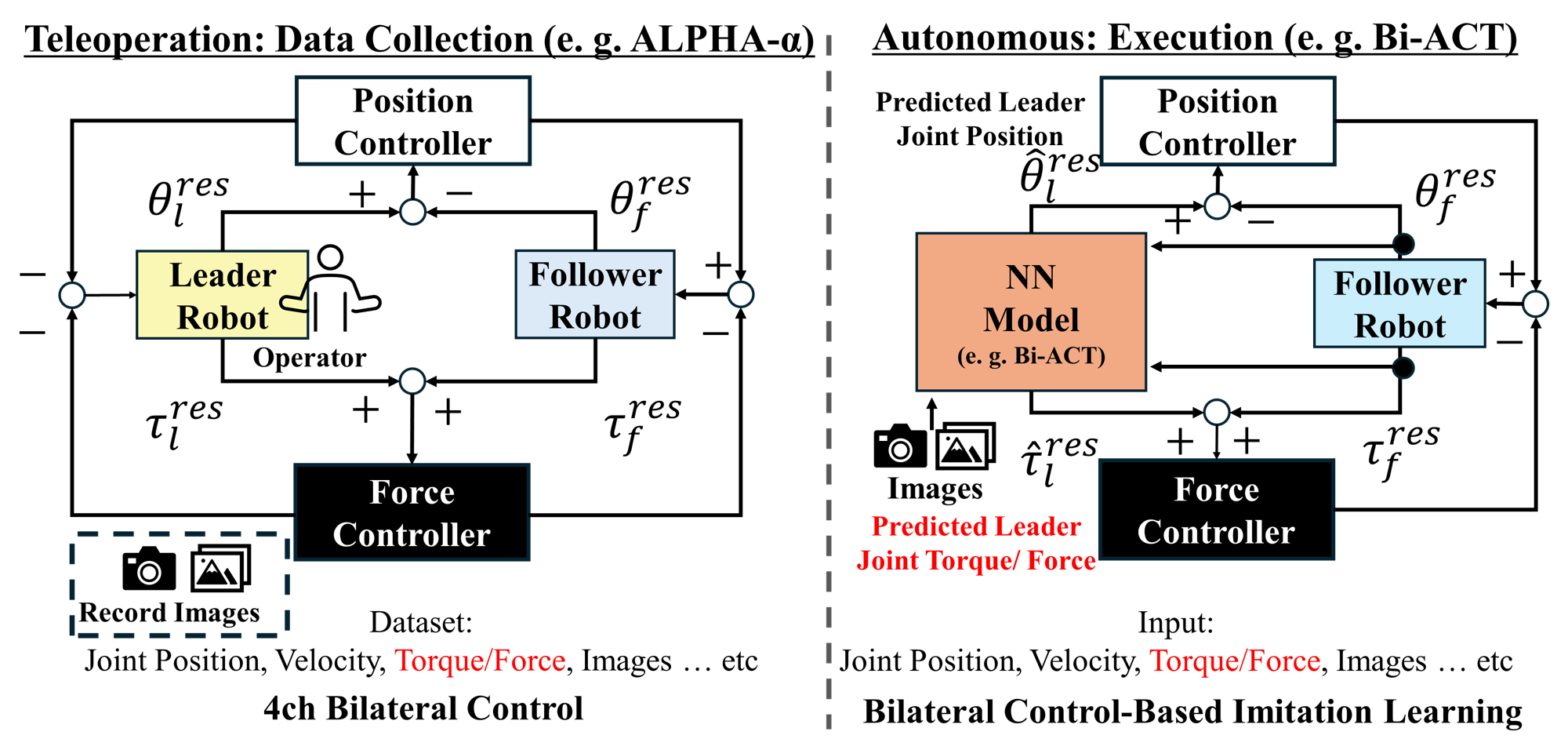}
   \caption{Bilateral Control-based Imitation Learning}
   \label{fig:bi}
\end{figure}

Bilateral control addresses this limitation by exchanging both position and force, enabling demonstrations rich in haptic and visual cues and yielding more robust generalization~\cite{buamanee2024biactbilateralcontrolbasedimitation}, as shown in Fig.~\ref{fig:bi}.
Among bilateral IL frameworks, Bi-ACT~\cite{buamanee2024biactbilateralcontrolbasedimitation} plays a central role by directly extending ACT to bilateral settings, offering a strong backbone for force-sensitive visuomotor learning. 
Recent extensions—including low-cost bilateral data collection~\cite{kobayashi2025alpha}, 
data augmentation~\cite{yamane20024data, kobayashi2024dabievaluationdataaugmentation}, 
the fast bilateral teleoperation and accurate dynamics model for Bi-ACT~\cite{yamane2025fastbilateralteleoperationimitation}, 
and Mamba-based motion encoders~\cite{tsuji2025manba}—demonstrate the flexibility and effectiveness of the bilateral IL paradigm. 
However, all such work assumes visually stable land environments and does not address the severe lighting variability characteristic and underwater settings.

A key gap therefore remains.
To the best of our knowledge, there exists no imitation learning framework that (i) brings bilateral control-based imitation learning (Bi-IL) to underwater robot manipulation, nor (ii) explicitly models underwater lighting as a latent factor in force-sensitive visuomotor control.
Bi-AQUA addresses both limitations simultaneously.

To close this gap, we introduce Bi-AQUA, the first underwater Bi-IL framework that explicitly models lighting at multiple levels of the visuomotor pipeline, as shown in Fig.~\ref{fig:tsr}.
Bi-AQUA builds upon Bi-ACT--style action chunking~\cite{buamanee2024biactbilateralcontrolbasedimitation, zhao2023learning} and incorporates a hierarchical lighting adaptation mechanism:
(1) a label-free \emph{Lighting Encoder} that learns lighting representations implicitly through the imitation objective,
(2) FiLM-based visual feature modulation for lighting-aware perception, and 
(3) a \emph{lighting token} added to the transformer encoder input for lighting-adaptive action generation.
Together, these components enable both perception and control to remain stable under static and dynamically changing lighting while leveraging rich force cues from bilateral demonstrations.

Our contributions are as follows:
\begin{itemize}
    \item Bi-AQUA is a first bilateral control–based imitation learning framework for underwater robot arms.
    \item We propose a lighting-aware visuomotor policy that integrates a label-free, implicitly supervised Lighting Encoder, FiLM modulation, and a lighting token for adaptive control.
    \item We demonstrated real-world gains over baseline and generalization to unseen lighting conditions, novel objects, and visual disturbances in underwater environments.
\end{itemize}

\begin{table}[t]
\centering
\caption{Comparison of representative approaches.}
\label{tab:rw_comparison}
\small
\renewcommand{\arraystretch}{1.05}
\setlength{\tabcolsep}{4pt}
\resizebox{\linewidth}{!}{
\begin{tabular}{c c c c c c c c}
\toprule
\textbf{Method} &
\textbf{Domain} &
\shortstack{\textbf{RGB}\\\textbf{Vision}} &
\textbf{Underwater} &
\shortstack{\textbf{Bilateral}\\\textbf{Control}} &
\shortstack{\textbf{Robot}\\\textbf{Policy}} &
\shortstack{\textbf{Lighting}\\\textbf{Modeling}} \\
\midrule

SelfLUID-Net~\cite{Varghese2025Sea-InginLow-Light} &
Image Enhancement &
\cmark{} &
\cmark{} &
\xmark{} &
\xmark{} &
\cmark{} \\

DeepSeeColor~\cite{Jamieson2023DeepSeeColor} &
Image Enhancement &
\cmark{} &
\cmark{} &
\xmark{} &
\xmark{} &
\cmark{} \\

RoLight~\cite{jin2025physicallybasedlightinggenerationrobotic} &
Manipulation&
\cmark{} &
\xmark{} &
\xmark{} &
\cmark{} &
\cmark{} \\

Underwater Bilateral~\cite{motoi2018development,nishi2025mrubimixedrealitybasedunderwater} &
Teleoperation &
\xmark{} &
\cmark{} &
\cmark{} &
\xmark{} &
\xmark{} \\

Ocean One~\cite{oceanone2016khatib,oceanone2020brantner} &
Teleoperation &
\cmark{} &
\cmark{} &
\cmark{} &
\xmark{} &
\xmark{} \\

UIVNAV~\cite{lin2024UINAV} &
Navigation(IL) &
\cmark{} &
\cmark{} &
\xmark{} &
\cmark{} &
\xmark{} \\

AquaBot~\cite{liu2024selfimprovingautonomousunderwatermanipulation} &
Manipulation (IL/RL) &
\cmark{} &
\cmark{} &
\xmark{} &
\cmark{} &
\xmark{} \\

U0~\cite{gu2025usimu0visionlanguageactiondataset} &
Manipulation/Navigation (IL:sim) &
\cmark{} &
\cmark{} &
\xmark{} &
\cmark{} &
\xmark{} \\

Bi-IL(w/o vision)~\cite{adachi2018imitation, tsuji2025manba, pmlr-v305-oishi25a} &
Manipulation (IL) &
\xmark{} &
\xmark{} &
\cmark{} &
\cmark{} &
\xmark{} \\

ACT~\cite{zhao2023learningfinegrainedbimanualmanipulation} &
Manipulation (IL) &
\cmark{} &
\xmark{} &
\xmark{} &
\cmark{} &
\xmark{} \\

Bi-ACT~\cite{buamanee2024biactbilateralcontrolbasedimitation,kobayashi2025alpha,yamane2025fastbilateralteleoperationimitation} &
Manipulation (IL) &
\cmark{} &
\xmark{} &
\cmark{} &
\cmark{} &
\xmark{} \\

\textbf{Bi-AQUA (Ours)} &
\textbf{Manipulation (IL)} &
\textbf{\cmark{}} &
\textbf{\cmark{}} &
\textbf{\cmark{}} &
\textbf{\cmark{}} &
\textbf{\cmark{}} \\
\bottomrule
\end{tabular}
}
\end{table}

\section{Related Work}
\subsection{Underwater Vision and Manipulation}

Underwater perception is degraded by wavelength-dependent attenuation, scattering, turbidity, and non-uniform illumination~\cite{Yang2019AnIn-DepthSurveyofUnderwaterImageEnhancement}. Extensive research addresses image restoration through physics-based, learning-based, and hybrid approaches~\cite{Jamieson2023DeepSeeColor, Akkaynak2019Sea-Thru, Demir2025JointOptimizationinUnderwaterImageEnhancement}, as well as large-scale datasets and low-light correction~\cite{Varghese2025Sea-InginLow-Light}. However, these methods optimize image quality rather than closed-loop manipulation performance.

Autonomous underwater manipulation has evolved from teleoperation~\cite{nishi2025mrubimixedrealitybasedunderwater} toward RL and IL-based control for navigation and manipulation~\cite{lin2024UINAV, liu2024selfimprovingautonomousunderwatermanipulation, gu2025usimu0visionlanguageactiondataset}. Nevertheless, existing systems either rely purely on vision or lack integration of force feedback and explicit lighting modeling, limiting robustness under dynamic underwater illumination.

Therefore, Bi-AQUA shifts the focus from image enhancement to manipulation performance, integrating force-sensitive bilateral control with explicit lighting modeling for robust underwater operation.

\subsection{Bilateral Control-based Imitation Learning}

Unilateral teleoperation systems such as ALOHA and Mobile ALOHA~\cite{zhao2023learningfinegrainedbimanualmanipulation, aloha2team2024aloha2enhancedlowcost, fu2024mobilealohalearningbimanual} depend solely on visual feedback, which restricts robustness in contact-rich tasks. Bilateral control-based imitation learning (Bi-IL) addresses this limitation by exchanging both position and force signals between leader and follower robots~\cite{tsuji2025surveyimitationlearningcontactrich, tsuji2025manba, adachi2018imitation}. 

Recent works integrate images into Bi-IL, notably Bi-ACT~\cite{buamanee2024biactbilateralcontrolbasedimitation, kobayashi2025alpha}, but evaluations are limited to visually stable terrestrial environments. Underwater settings introduce severe photometric distortions, rendering proprioception-only approaches perceptually insufficient and vision-based Bi-IL vulnerable to lighting shifts. To our knowledge, no prior work unifies Bi-IL, real underwater deployment, and explicit lighting modeling within a single visuomotor policy.

Bi-AQUA extends bilateral imitation learning to real underwater environments by explicitly incorporating lighting-aware visual conditioning into a force-sensitive control framework.

\subsection{Robust Visual Policies and Lighting Modeling}

Visuomotor policies are highly sensitive to photometric shifts. Robustness is commonly improved through domain randomization~\cite{tobin2017Domainrandomizationfortransferringdeepneuralnetworks} or physically based relighting~\cite{jin2025physicallybasedlightinggenerationrobotic}, but these approaches treat illumination implicitly and do not provide an internal lighting representation within the policy.

Conditional modulation methods such as FiLM~\cite{perez2018film} enable feature adaptation~\cite{zhu2025equactse3equivariantmultitasktransformer}, yet lighting is typically not disentangled as a structured latent factor influencing both perception and action generation.

In contrast, Bi-AQUA encodes lighting as a label-free latent variable and integrates it hierarchically via FiLM modulation and a dedicated lighting token, enabling adaptation to rapidly changing underwater illumination within the control policy itself.

Taken together, prior work has addressed underwater perception, bilateral imitation learning, and visual robustness largely in isolation. Bi-AQUA bridges these directions by unifying force-sensitive bilateral control with explicit, hierarchical lighting modeling within a single visuomotor policy for real underwater manipulation.
\begin{figure*}[t]
  \centering
   \includegraphics[width=\linewidth]{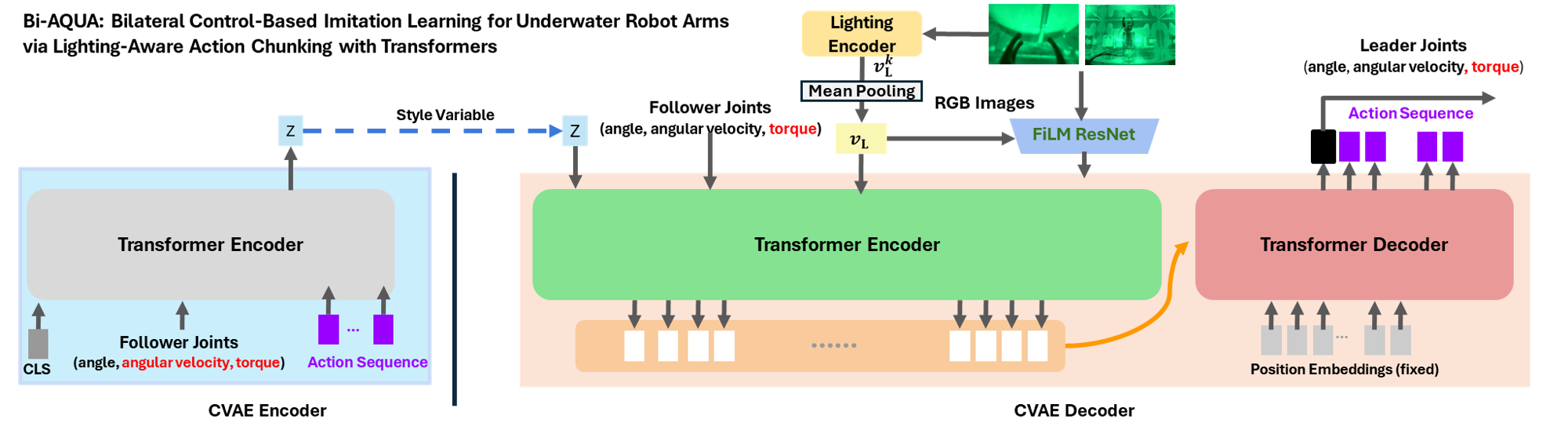}
   \caption{Overview of Bi-AQUA. Given multi-view underwater observations and follower joint states, Bi-AQUA extracts lighting-aware visual features, fuses them with proprioception, and predicts leader-side action chunks within a bilateral control loop.}
   \label{fig:model}
\end{figure*}
\section{Bi-AQUA Method}

\subsection{Overview}
Bi-AQUA inherits Bi-ACT's leader--follower bilateral control, leveraging both position and force information, and introduces lighting-aware visual processing at three levels: (i) a \emph{Lighting Encoder} that extracts compact lighting embeddings from RGB images, (ii) \emph{FiLM-based feature-wise modulation} of backbone features conditioned on the lighting embedding, and (iii) an explicit \emph{lighting token} added to the transformer encoder input, which flows through the encoder to the decoder for sequence-level conditioning. This hierarchical design enables the policy to adapt to diverse and time-varying underwater lighting while preserving force-sensitive manipulation.

\subsection{Data Collection}

\begin{figure}[t]
  \centering
   \includegraphics[width=\linewidth]{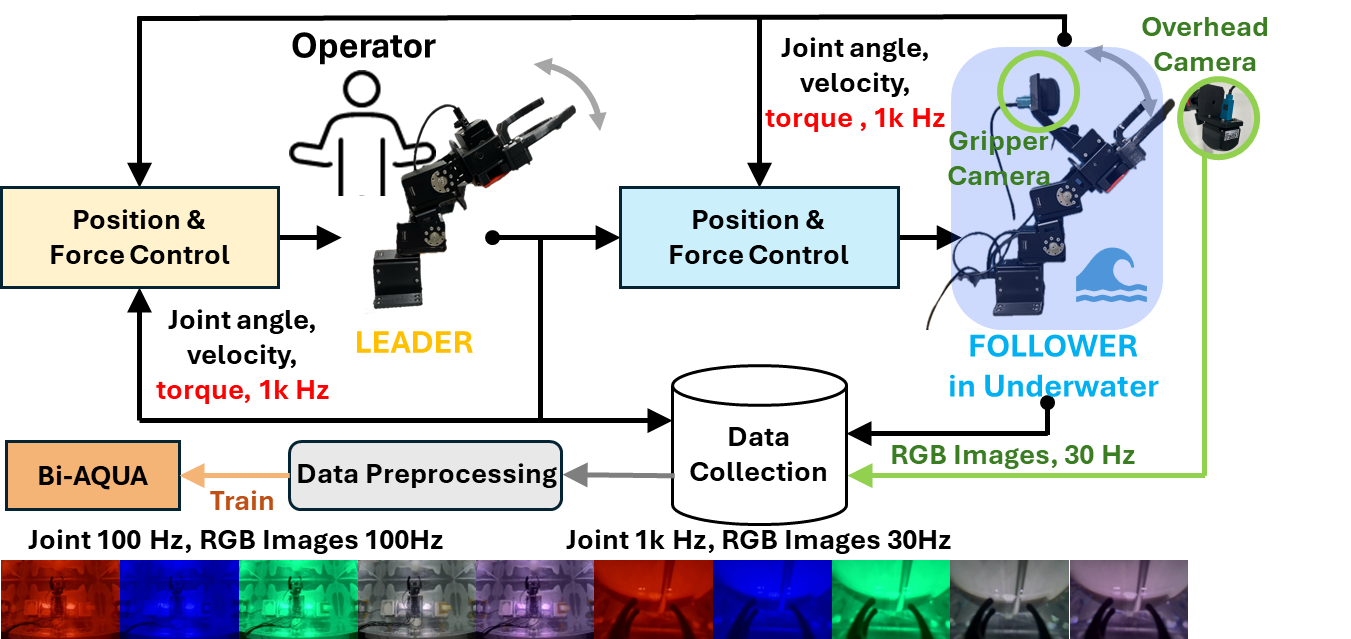}
   \caption{Data collection of Bi-AQUA.}
   \label{fig:data_collection}
\end{figure}

Data collection follows the bilateral control paradigm of Bi-ACT, adapted to an underwater follower as shown in Fig.~\ref{fig:data_collection}. A human operator manipulates a leader robot in air, and a follower robot executes mirrored motions underwater.

Bilateral control enforces simultaneous position tracking and force feedback between leader and follower according to
\begin{equation}
    \theta_l - \theta_f = 0, \quad \tau_l + \tau_f = 0,
\end{equation}
where $\theta$ and $\tau$ denote joint position and torques, respectively. Position is obtained from encoders, velocities by differentiation, and torques by a disturbance observer (DOB)~\cite{ohnishi1996motion} and a reaction force observer (RFOB)~\cite{murakami2002torque}, without dedicated force sensors.

We record multi-modal observations from the underwater follower: RGB images from $N_c$ cameras at frequency $f_{\text{img}}$, follower joint data (position, velocity, torque) at $f_{\text{joint}}$, and synchronized leader–follower joint states $\mathbf{l}_t, \mathbf{f}_t \in \mathbb{R}^{3N}$.
All streams are temporally standardized to the training frequency $f_{\text{train}}$: images are aligned by selecting the nearest frame to each training timestamp, and joint data are downsampled by taking every $\lfloor f_{\text{joint}} / f_{\text{train}} \rfloor$ sample. Images are resized to $H_{\text{img}} \times W_{\text{img}}$ via letterboxing. After preprocessing, all modalities are aligned at $f_{\text{train}}$ and stored in a unified dataset.

To ensure robustness, demonstrations are collected under diverse lighting spectra and intensities. Bilateral control allows the operator to sense contact forces, enabling safe and precise underwater manipulation during teleoperation.

\subsection{Learning Model}

Bi-AQUA extends the Bi-ACT architecture by incorporating lighting-aware visual processing into a transformer-based action prediction model, as shown in Fig.~\ref{fig:model}.
The system consists of:
\begin{enumerate}
    \item a \emph{ label-free lighting-aware visual encoder},
    \item a \emph{joint state encoder},
    \item a \emph{transformer-based action prediction module},
    \item \emph{bilateral control integration} for closed-loop execution.
\end{enumerate}

\subsubsection{Lighting-Aware Visual Processing}

Given multi-view observations $\mathbf{I} = \{\mathbf{I}_i\}_{i=1}^{N_c}$, our goal is to extract manipulation-relevant features while explicitly modeling lighting. We introduce three components: a Lighting Encoder, a FiLM-modulated backbone, and a lighting token.

\begin{figure}[t]
  \centering
   \includegraphics[width=\linewidth]{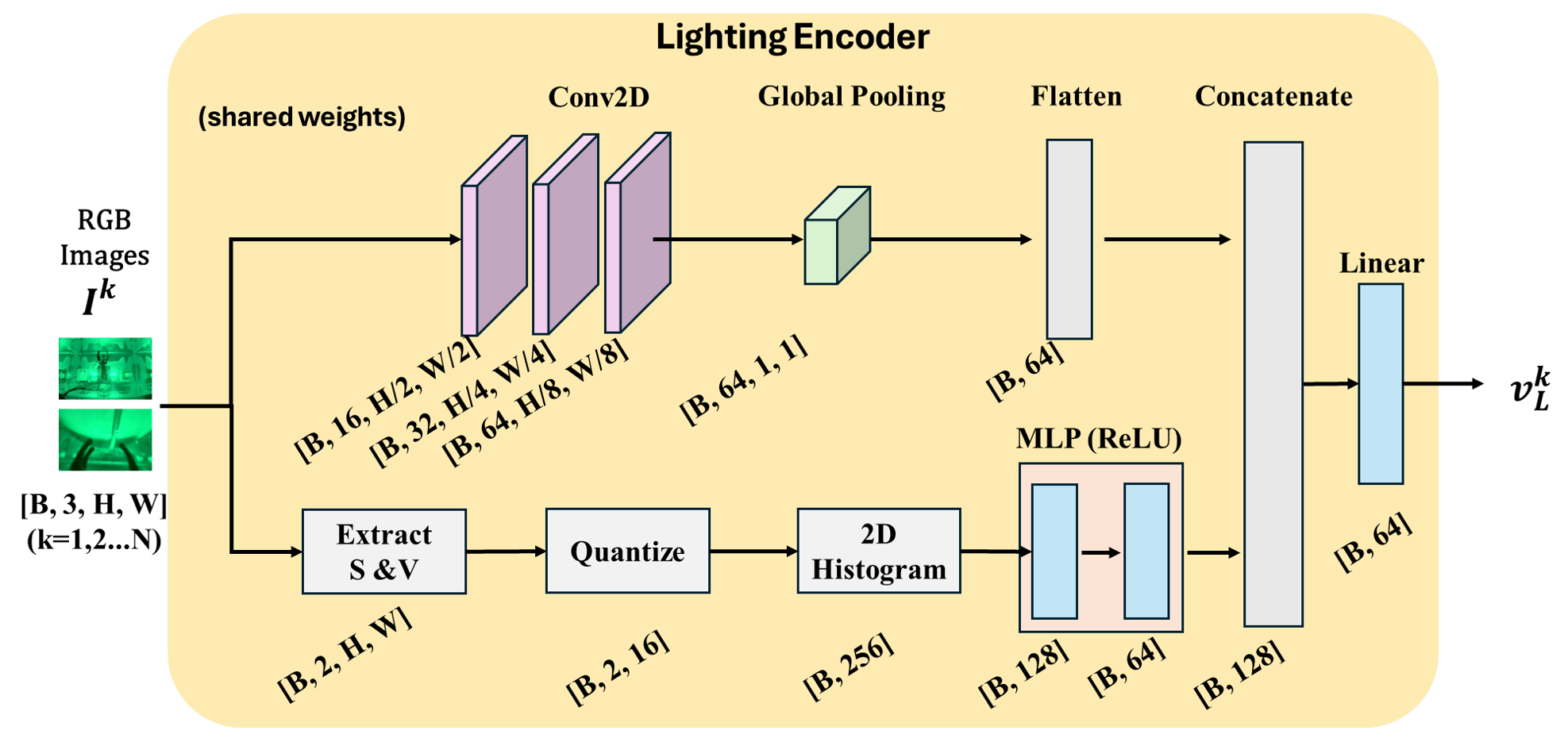}
   \caption{Lighting Encoder. A convolutional path captures spatial lighting cues, while a histogram path models saturation--value statistics. The two paths are fused into a compact lighting representation without any lighting annotations.}
   \label{fig:le}
\end{figure}

\paragraph{Lighting Encoder.}
As shown in Fig.~\ref{fig:le}, we design a Lighting Encoder $\mathcal{E}_L: \mathbb{R}^{3 \times H \times W} \rightarrow \mathbb{R}^{d_L}$ that produces a compact lighting representation $\mathbf{v}_L \in \mathbb{R}^{d_L}$ from RGB images \emph{without} lighting or color annotations, which are difficult to define in underwater settings.

The encoder adopts a dual-path architecture. A \emph{convolutional path} processes the input through $L_{\text{conv}}$ convolutional layers with channels $3 \to C_1 \to C_2 \to C_3$, followed by ReLU and global average pooling to yield spatial features $\mathbf{f}_{\text{conv}} \in \mathbb{R}^{C_3}$. A \emph{histogram path} computes a 2D histogram $\mathbf{H} \in \mathbb{R}^{B_{\text{hist}}^2}$ over saturation and value (SV) channels and passes it through a two-layer MLP to obtain $\mathbf{f}_{\text{hist}} \in \mathbb{R}^{d_{\text{hist}}}$. The final lighting embedding is
\begin{equation}
\mathbf{v}_L = \mathbf{W}_L [\mathbf{f}_{\text{conv}}; \mathbf{f}_{\text{hist}}] \in \mathbb{R}^{d_L},
\end{equation}
with $\mathbf{W}_L \in \mathbb{R}^{d_L \times (C_3 + d_{\text{hist}})}$ and concatenation $[\cdot;\cdot]$. For multiple cameras, we use a shared encoder and average across views:
\begin{equation}
\mathbf{v}_L = \frac{1}{N_c} \sum_{i=1}^{N_c} \mathcal{E}_L(\mathbf{I}_i).
\end{equation}

\paragraph{FiLM-modulated visual backbone.}
Following the formulation of FiLM~\cite{perez2018film}, we modulate each convolutional feature map via a feature-wise affine transformation whose parameters are generated from the lighting embedding.
For a feature map $\mathbf{F}_\ell \in \mathbb{R}^{B \times C_\ell \times H_\ell \times W_\ell}$ at layer $\ell$, the FiLM operation is
\begin{equation}
\operatorname{FiLM}(\mathbf{F}_\ell \mid \boldsymbol{\gamma}_\ell, \boldsymbol{\beta}_\ell)
=
(1 + \boldsymbol{\gamma}_\ell) \odot \mathbf{F}_\ell + \boldsymbol{\beta}_\ell,
\end{equation}
where $\boldsymbol{\gamma}_\ell, \boldsymbol{\beta}_\ell \in \mathbb{R}^{C_\ell}$ are per-channel scaling and shifting coefficients, broadcast spatially over $(H_\ell, W_\ell)$, and $\odot$ denotes element-wise multiplication.
This formulation preserves the identity mapping when $\boldsymbol{\gamma}_\ell = \mathbf{0}$ and $\boldsymbol{\beta}_\ell = \mathbf{0}$, facilitating stable training initialization.

Both parameters are produced by learned functions of the conditioning input---here, the lighting vector $\mathbf{v}_L$:
\begin{equation}
\boldsymbol{\gamma}_\ell = f_\ell(\mathbf{v}_L),
\qquad
\boldsymbol{\beta}_\ell = h_\ell(\mathbf{v}_L),
\end{equation}
where $f_\ell$ and $h_\ell$ are implemented as linear projections
\begin{equation}
f_\ell(\mathbf{v}_L) = \mathbf{W}_{\gamma}^{(\ell)} \mathbf{v}_L,
\qquad
h_\ell(\mathbf{v}_L) = \mathbf{W}_{\beta}^{(\ell)} \mathbf{v}_L,
\end{equation}
with $\mathbf{W}_{\gamma}^{(\ell)}, \mathbf{W}_{\beta}^{(\ell)} \in \mathbb{R}^{C_\ell \times d_L}$.
In our implementation, FiLM is applied to the final ResNet layer.
The FiLM-modulated features are then flattened into tokens, augmented with 2D sinusoidal positional embeddings, and concatenated across camera views to form the visual token sequence $\mathbf{F}_{\text{vis}} \in \mathbb{R}^{N_{\text{tokens}} \times d_h}$ used by the transformer decoder.

\subsubsection{Training Objective}
Bi-AQUA employs a Conditional Variational Autoencoder (CVAE) with a transformer backbone, following Bi-ACT while augmenting it with lighting-aware conditioning. Given follower state $\mathbf{f}_t$ and visual observations $\mathbf{I}_t$, the policy predicts a chunk of leader actions $\hat{\mathbf{l}}_{t:t+k}$.
The lighting token is added to the transformer encoder input alongside latent and proprioceptive tokens. The encoder processes these tokens together, and the resulting memory (which includes the lighting information) is then accessed by the transformer decoder via cross-attention, enabling the decoder to adapt action generation to the current underwater lighting.

We train Bi-AQUA end-to-end using behavior cloning with a CVAE-style latent space:
\begin{equation}
\mathcal{L} = \mathcal{L}_{\text{action}} + \lambda_{\text{KL}} \mathcal{D}_{\text{KL}},
\end{equation}
where $\mathcal{L}_{\text{action}}$ is an L1 loss between predicted and ground-truth leader actions, and $\mathcal{D}_{\text{KL}}$ is the KL divergence between the learned latent distribution and a standard normal prior, weighted by $\lambda_{\text{KL}}$. The Lighting Encoder, FiLM layers, and transformer modules are optimized jointly.
\begin{figure}[t]
  \centering
   \includegraphics[width=\linewidth]{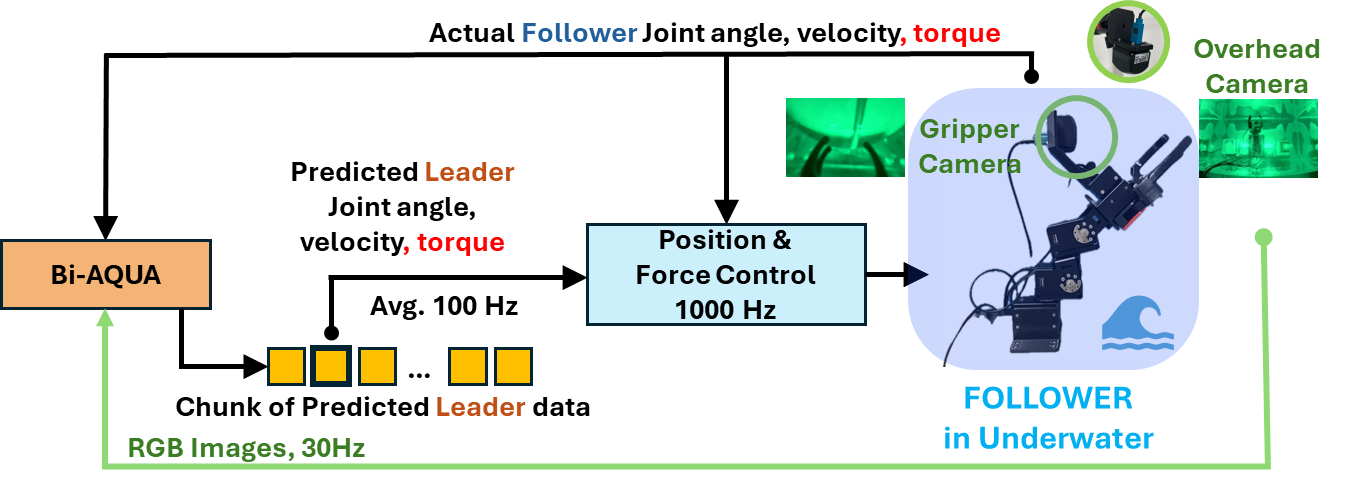}
   \caption{Inference pipeline of Bi-AQUA.}
   \label{fig:inferrence}
\end{figure}
\subsection{Inference}
At test time, Bi-AQUA runs in closed loop with the bilateral controller, as shown in Fig.~\ref{fig:inferrence}. At each inference step, the model receives current multi-view images and follower joint state, computes $\mathbf{v}_L$ with the Lighting Encoder, applies FiLM modulation to the backbone, samples a latent action code from the prior, and uses the transformer decoder to generate an action chunk for the leader robot.

The predicted leader actions are converted into follower commands by the bilateral controller and executed at control frequency $f_{\text{control}}$. Since the policy operates at a lower inference rate $f_{\text{inference}}$, we maintain an action chunk buffer of length $k$ and output actions at $f_{\text{train}}$ between inference calls, ensuring smooth and responsive motion.

The combination of Lighting Encoder, FiLM-based visual modulation, and a lighting token yields a lighting-aware visuomotor policy that maintains reliable underwater manipulation even under challenging and dynamically changing lighting.

\section{Experiments}
\subsection{Experimental Setup}

\begin{figure}[t]
    \centering
    \includegraphics[width=\linewidth]{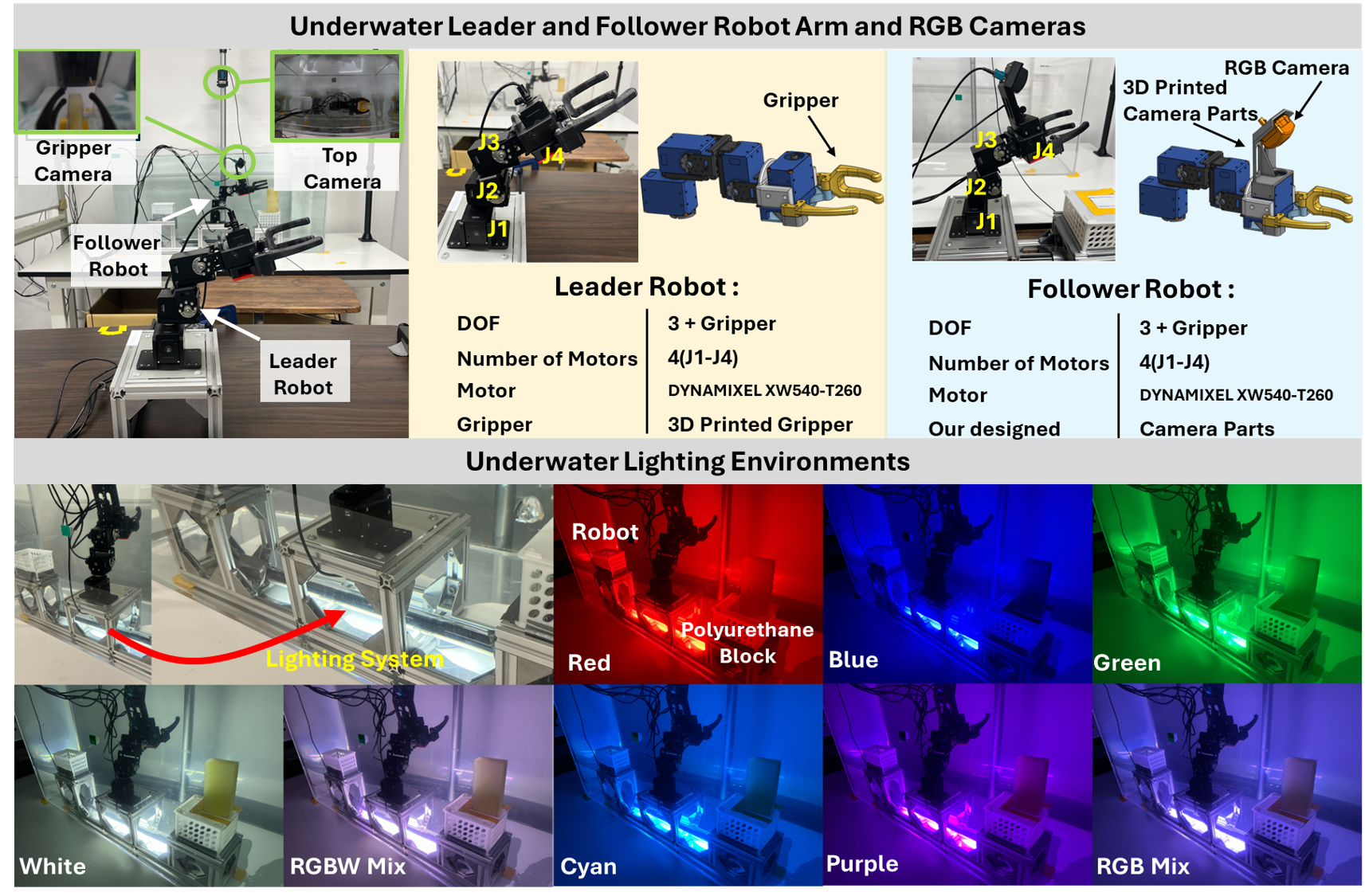}
    \caption{Experimental Environments}
    \label{fig:experimental_setup}
\end{figure}

We evaluate Bi-AQUA on a bilateral leader--follower system shown in Fig.~\ref{fig:experimental_setup}.
Each arm has three revolute joints (J1--J3) and a gripper (J4), actuated by Dynamixel XW540-T260 (IP68) servo motors.
The follower arm is mounted on a 200\,mm pedestal inside a water tank of $900 \times 450 \times 450$\,mm.

To create reproducible underwater lighting variation, we install an RGBW LED aquarium light (FEDOUR RD10-400RGBW, 38\,cm, 7\,W, 21 LEDs, IP68) beneath the follower pedestal and use eight lighting modes:
\emph{red, blue, green, white, rgbw, cyan, purple}, and \emph{changing}.
All static modes correspond to the manufacturer’s 100\% brightness presets, while the dynamic \emph{changing} mode uses the built-in “M5” program, which cycles through colors every 2\,s.
This setup enables controlled evaluation under both static and rapidly varying illumination.

Bi-AQUA receives RGB observations from two cameras (TIER IV C1 120): a gripper-mounted camera for local close-up views and a top camera observing the global workspace.
\subsection{Tasks and Evaluation Protocol}
\label{sec:tasks_protocol}

\subsubsection{Common Protocol}
Across all tasks, we use a unified train/test lighting protocol to isolate the effect of lighting variation.
Unless otherwise noted, we collect bilateral teleoperation demonstrations under five training lighting modes
(\emph{red, blue, green, white, rgbw}) and evaluate under all eight modes, including two unseen colors
(\emph{cyan, purple}) and the dynamic \emph{changing} condition.
Each test configuration is evaluated over 5 autonomous rollouts, and we report success rate (\%).

We evaluate three manipulation tasks with progressively increasing temporal horizon and contact complexity:
\begin{itemize}
    \item \textbf{Pick-and-Place Task}: 
    a lighting-robustness benchmark in which the robot transports an object across the tank.
    
    \item \textbf{Drawer Closing Task}: 
    a long-horizon sequential task with an additional drawer-closing operation.
    
    \item \textbf{Peg Extraction Task}: 
    a contact-rich precision manipulation task requiring stable grasping and controlled disengagement.
\end{itemize}

\subsubsection{Pick-and-Place Task}
\label{sec:task1}
\begin{figure}[t]
  \centering
  \includegraphics[width=\linewidth]{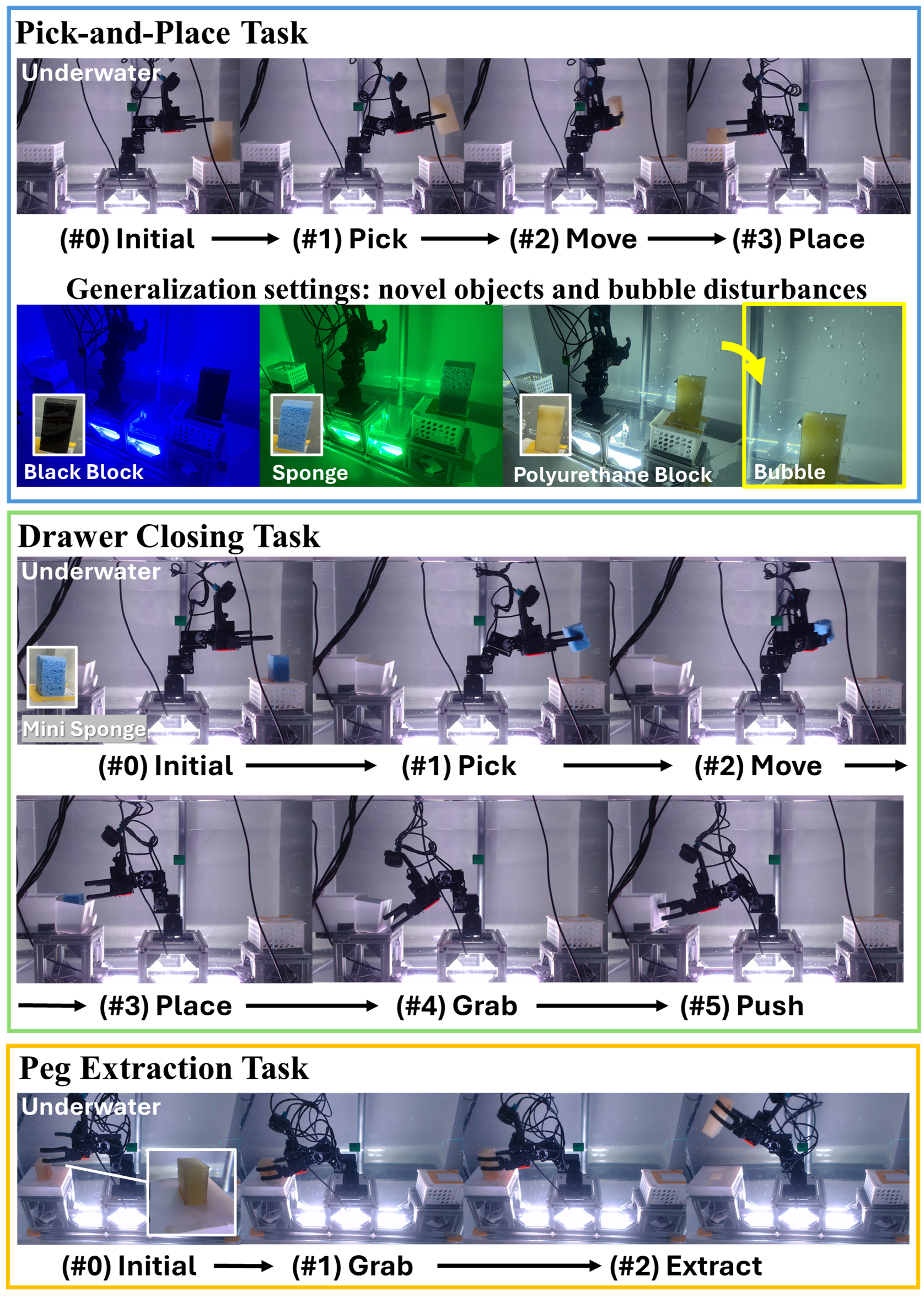}
  \caption{Overview of underwater manipulation tasks}
  \label{fig:task_overview}
\end{figure}

Pick-and-Place task is designed to directly stress lighting robustness in underwater manipulation policy.
The follower must grasp the polyurethane block on the right side of the tank and place it into a basket on the opposite side,
passing through four stages:
\#0 initial pose, \#1 pick up, \#2 transport, \#3 place.
A rollout is successful if the object ends inside the basket without being dropped.

To test robustness beyond the training object and scene, we additionally evaluate
(a) a black rubber block, (b) a blue sponge, and (c) bubbles around the polyurethane block (Fig.~\ref{fig:task_overview}).
These settings are evaluated under the same eight lighting modes with the same success criterion.

\subsubsection{Drawer Closing Task}
\label{sec:task2}

Drawer closing task extends Pick-and-Place task into a long-horizon sequential task by appending a drawer closing operation.
After placing the object into the target container, the robot approaches a drawer and performs a closing contact-rich motion.
We define \emph{full-task success} if both conditions hold:
\textbf{(i)} the object is placed in the target container and \textbf{(ii)} the drawer reaches the closed state.

\subsubsection{Peg Extraction Task}
\label{sec:task3}
Peg Extraction evaluates contact-rich precision manipulation under severe underwater photometric variation.  
At the beginning of each rollout, the peg is partially inserted into a mating hole. 
The robot must establish a stable grasp and extract the peg without losing contact stability.
The radial clearance between the peg and the hole is 1.5\,mm, resulting in tight geometric tolerance and substantial frictional contact along the insertion surface. 
Consequently, successful extraction requires careful modulation of grasp force and pulling motion to overcome friction while avoiding excessive force that may destabilize the grasp.
A rollout is considered successful if the peg is completely removed from the hole and securely held by the robot at the end of the episode.

\begin{figure}[t]
  \centering
  \includegraphics[width=\linewidth]{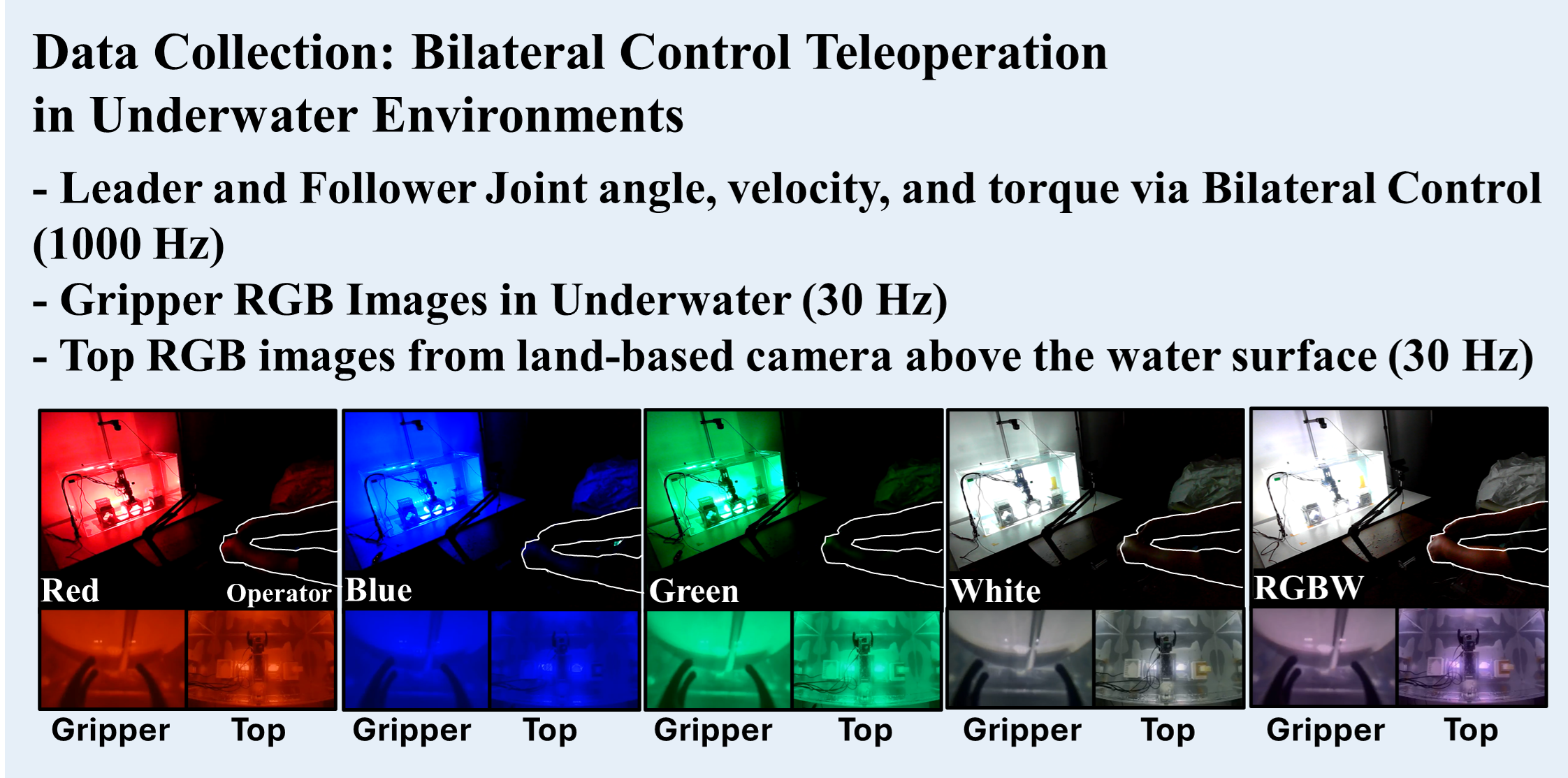}
  \caption{Bilateral teleoperation data collection under five training lighting modes.}
  \label{fig:ex_data_collection}
\end{figure}

\subsection{Dataset and Training Configuration}

\subsubsection{Demonstration Collection and Preprocessing}
We use the bilateral control data-collection pipeline for all tasks.
The policy is trained and evaluated only from follower-side observations, while leader joint states are used as action supervision.

For each task, we collect 10 bilateral teleoperation episodes (two per training lighting mode) under five training conditions:
\emph{red, blue, green, white}, and \emph{rgbw}.
For Pick-and-Place, demonstrations follow a consistent motion pattern so that the main variation comes from lighting rather than operator strategy.

Two RGB cameras (top and gripper views) record at 30\,Hz with resolution $1920 \times 1280$, and images are resized to $480 \times 320$.
Joint position, velocity, and torque for $N=4$ joints are recorded from both arms at 1{,}000\,Hz.
For training, joint data are downsampled to 100\,Hz, and images are aligned to the same rate by selecting the nearest frame at each 10\,ms timestamp.
Each training sample therefore consists of two RGB views, the follower joint state, and the corresponding leader joint state.

Although demonstrations are collected under known lighting modes, no lighting IDs or manual labels are provided.
The Lighting Encoder is trained end-to-end through the imitation objective and learns a latent representation directly from raw images.

\subsubsection{Compared Methods and Ablations (Pick-and-Place task)}

\begin{table}[t]
    \centering
    \caption{Ablation of lighting-aware components.}
    \label{tab:ablation}
    \small
    \setlength{\tabcolsep}{2pt}
    \begin{tabular}{lccc}
      \toprule
      Method & Lighting Encoder & Lighting Token & FiLM \\
      \midrule
      Bi-ACT (Baseline)       & --    & --    & --    \\
      Bi-ACT+LE-Token         & \cmark & \cmark & --    \\
      Bi-ACT+LE-FiLM          & \cmark & --    & \cmark \\
      \textbf{Bi-AQUA (Ours)} & \cmark & \cmark & \cmark \\
      \bottomrule
    \end{tabular}
\end{table}

All variants share the same Bi-ACT-style visuomotor backbone and differ only in three lighting-aware components:
(i) a 64-D Lighting Encoder, (ii) FiLM modulation in the visual backbone, and (iii) a lighting token added to the transformer encoder input.
Baseline and ablation comparisons are conducted on Pick-and-Place to isolate the effect of explicit lighting modeling, while Drawer Closing and Peg Extraction are used to evaluate task generality for Bi-AQUA.

\subsubsection{Training Configuration}
All models are trained offline with the same preprocessing and optimization settings.
We use a Bi-ACT-style CVAE action-chunking policy with KL regularization, optimized by AdamW with learning rate $1 \times 10^{-5}$ for both the ResNet-18 backbone and the remaining modules.
The KL weight is set to $\lambda_{\text{KL}} = 1.0$.
The transformer uses 8 attention heads, hidden dimension 512, and feedforward dimension 3{,}200.
\begin{figure}[t]
  \centering
   \includegraphics[width=\linewidth]{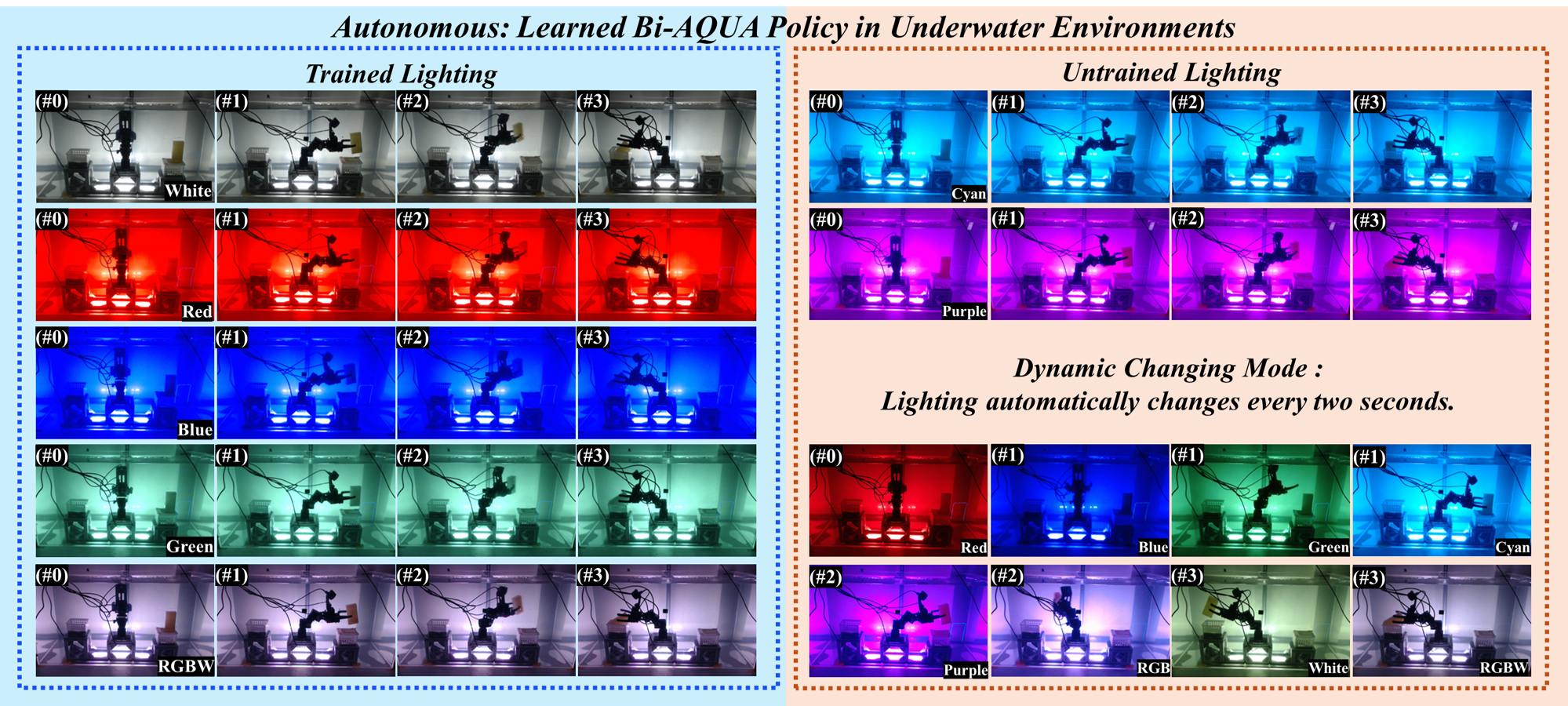}
   \caption{Execution of Bi-AQUA under lighting conditions.}
   \label{fig:ex_auto}
\end{figure}

\subsection{Experimental Results}
\subsubsection{Pick-and-Place task: Lighting Robustness and Ablation}

\begin{table}[t]
  \centering
  \caption{
  Task success rate (\%) under different lighting conditions.
  }
  \resizebox{\linewidth}{!}{
  \begin{tabular}{l c c c c c c c c}
  \toprule
  \multirow{2}{*}{\textbf{Method}} 
    & \multicolumn{5}{c}{\textbf{Trained Lighting}} 
    & \multicolumn{3}{c}{\textbf{Untrained Lighting}} \\
  \cmidrule(lr){2-6} \cmidrule(lr){7-9}
    & red & blue & green & white & rgbw & cyan & purple & changing \\
  \midrule
  Bi-ACT (Baseline)       & 20  & 0  & 0   & 100 & 0   & 0   & 0   & 0   \\
  Bi-ACT+LE-Token         & 0   & 0  & 0   & 0   & 0   & 0   & 0   & 0   \\
  Bi-ACT+LE-FiLM          & 80  & 100& 100 & 100 & 100 & 60  & 100 & 20  \\
  \textbf{Bi-AQUA (Ours)} & \textbf{100} & \textbf{80} & \textbf{100} & \textbf{100} & \textbf{100} & \textbf{100} & \textbf{100} & \textbf{100} \\
  \midrule
  \multicolumn{9}{c}{\textbf{Bi-AQUA generalization to novel objects and disturbances}} \\
  \midrule
  Black Block & 60 & 0  & 60  & 100 & 100 & 100 & 100 & 60 \\
  Sponge      & 100& 20 & 100 & 20  & 0   & 80  & 20  & 40 \\
  Bubble      & 20 & 80 & 80  & 100 & 80  & 0   & 100 & 40 \\
  \bottomrule
  \end{tabular}
  }
  \label{tab:lighting_results_all}
\end{table}

\begin{figure}[t]
  \centering
  \includegraphics[width=\linewidth]{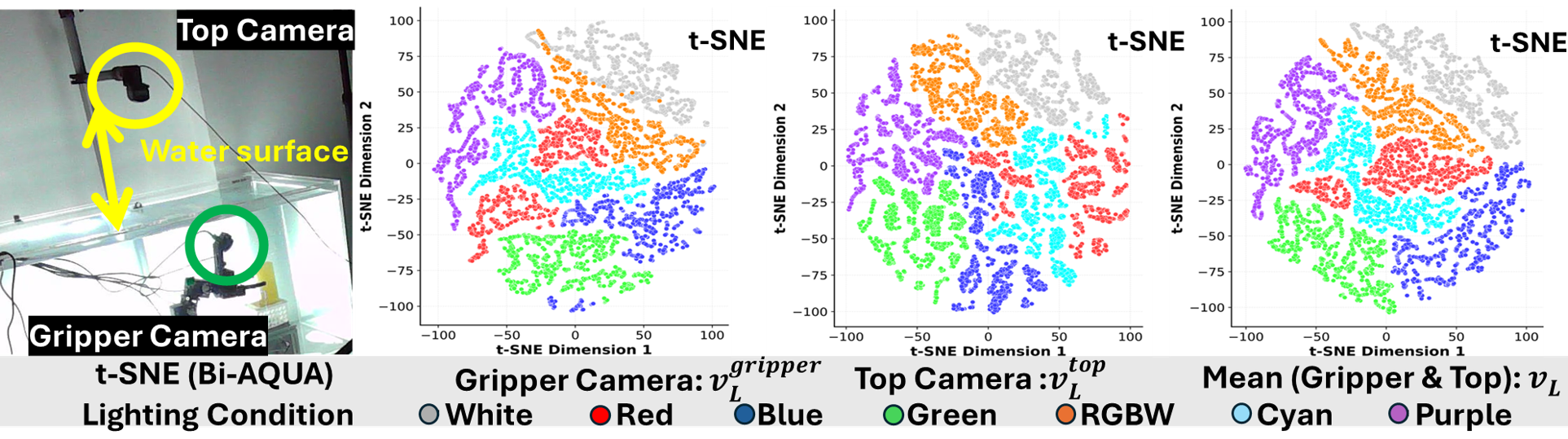}
  \caption{t-SNE of each camera.}
  \label{fig:tsne}
\end{figure}

\begin{figure}[t]
  \centering
  \includegraphics[width=\linewidth]{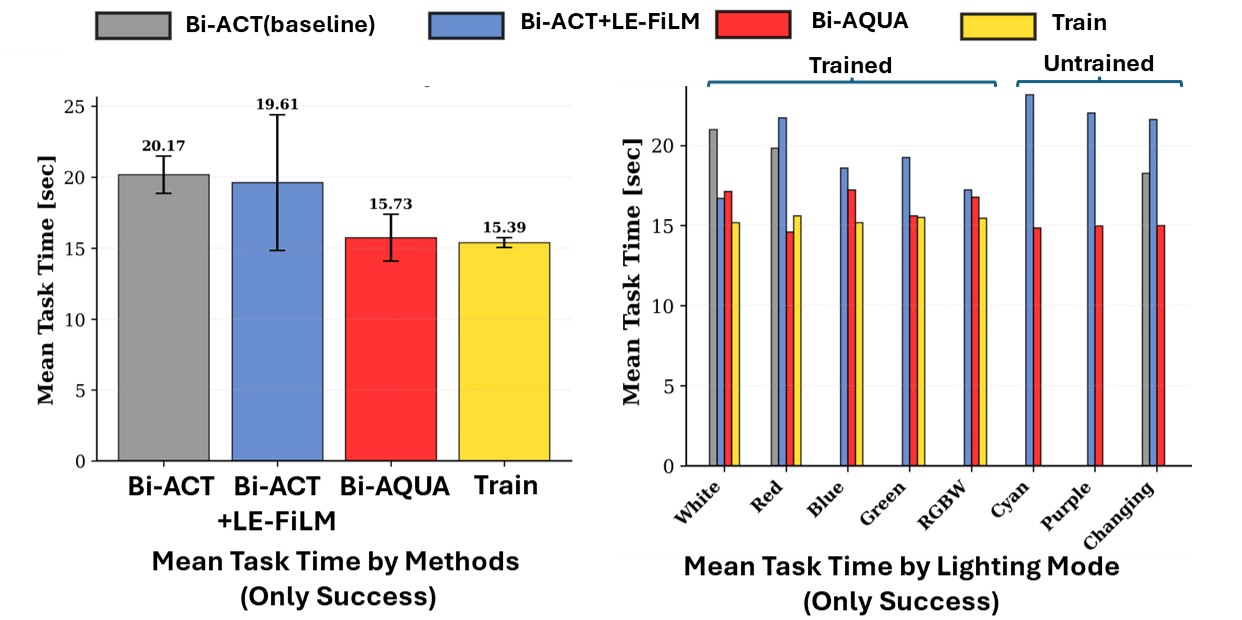}
  \caption{Execution time profiles (Pick-and-Place task).}
  \label{fig:Time}
\end{figure}

\begin{table}[t]
\centering
\scriptsize
\caption{Success rate (\%) across lighting and multi-stage task}
\resizebox{\linewidth}{!}{
\begin{tabular}{lcccccccc}
\toprule
  \multirow{2}{*}{\textbf{Method}} 
    & \multicolumn{5}{c}{\textbf{Trained Lighting}} 
    & \multicolumn{3}{c}{\textbf{Untrained Lighting}} \\
  \cmidrule(lr){2-6} \cmidrule(lr){7-9}
 & red & blue & green & white & rgbw & cyan & purple & changing \\
\midrule
\multicolumn{9}{c}{\textbf{Multi-stage drawer closing contact-rich task.}} \\
Bi-AQUA (\textbf{w/o Force}) & 40 & 40 & 100 & 60 & 40 & 100 & 100 & 40 \\
Bi-AQUA (Ours)    & 100 & 100 & 100 & 80 & 80 & 100 & 80 & 80 \\
\multicolumn{9}{c}{\textbf{Peg-out-Hole contact-rich task.}} \\
Bi-AQUA (\textbf{w/o Force}) & 0 & 40 & 60 & 60 & 80 & 40 & 0 & 0 \\
Bi-AQUA (Ours)    & 100 & 100 & 100 & 100 & 100 & 100 & 80 & 100 \\
\bottomrule
\end{tabular}}

\label{tab:all_results}
\end{table}

Table~\ref{tab:lighting_results_all} reports Pick-and-Place success rates under all eight lighting modes, including two unseen colors
(\emph{cyan} and \emph{purple}) and the dynamic \emph{changing} condition.
Bi-AQUA shows strong robustness, achieving \emph{100\%} success in seven of the eight modes and maintaining \emph{80\%} under blue illumination, which is particularly challenging because of wavelength-dependent attenuation and reduced underwater contrast.
Notably, Bi-AQUA also achieves \emph{100\%} success in the \emph{changing} mode, where the illumination switches every 2\,s and scene appearance changes substantially within a single action horizon.
These results indicate robustness to both static photometric shifts and temporally varying illumination.

In contrast, the lighting-agnostic Bi-ACT baseline succeeds only under white illumination (100\%) and partially under red illumination (20\%), while failing in all remaining conditions despite using the same demonstrations.
This gap shows that explicit lighting modeling is essential for reliable underwater visuomotor control.

The ablation results in Table~\ref{tab:ablation} further clarify the role of each component.
\emph{Bi-ACT+LE-Token} fails across all modes, indicating that sequence-level conditioning alone cannot compensate for underwater color and contrast distortion.
\emph{Bi-ACT+LE-FiLM} performs strongly in several static and unseen lighting conditions, confirming the importance of lighting-aware visual feature modulation, but drops sharply in the dynamic \emph{changing} mode.
Only the full Bi-AQUA model, combining the Lighting Encoder, FiLM modulation, and the lighting token, maintains uniformly high success across all conditions, demonstrating that hierarchical lighting adaptation is necessary for robust underwater manipulation.

This interpretation is supported by the t-SNE visualization in Fig.~\ref{fig:tsne}.
The learned lighting embeddings form meaningful clusters by lighting condition, especially for the gripper camera and the fused multi-view representation.
The top-camera embedding shows larger overlap, likely due to water-surface reflection and global scene ambiguity, whereas averaging the two camera embeddings yields clearer separation, including for unseen lighting conditions.
These results suggest that the Lighting Encoder learns a structured latent representation of underwater illumination without explicit labels, and that multi-view fusion improves lighting discriminability.

Fig.~\ref{fig:Time} shows the execution time analysis.
Averaged over all lighting conditions, Bi-AQUA completes the task in $15.73$\,s, closely matching human bilateral teleoperation at $15.39$\,s, while the baselines are slower (Bi-ACT: $20.17$\,s; Bi-ACT+LE-FiLM: $19.61$\,s).
Thus, the proposed lighting-aware design improves robustness without sacrificing execution efficiency, enabling smooth and efficient task execution under severe underwater photometric variation.

\subsubsection{Pick-and-Place task: Generalization to Novel Objects and Disturbances}
To evaluate whether Bi-AQUA overfits to the polyurethane block used during training, we test three additional settings:
a black rubber block, a blue sponge, and bubbles surrounding the manipulated object (Fig.~\ref{fig:task_overview}).
Table~\ref{tab:lighting_results_all} shows that Bi-AQUA generalizes beyond the training distribution, maintaining high success on the black block (60--100\%)
and non-trivial performance under severe appearance shifts (blue sponge) and compounded visual distortions (bubbles).
These results indicate robustness not only to lighting variation but also to object and environmental changes without additional training.

\subsubsection{Drawer closing task: Long-Horizon drawer Closing}
In the multi-stage drawer closing task, Bi-AQUA consistently outperforms the force-ablated variant across both seen and unseen lighting conditions. 
Under trained lighting modes, the full model achieves success rates of 100\% in red, blue, and green illumination, and 80\% in white and rgbw. 
In contrast, the force-ablated policy exhibits substantial degradation, with success rates dropping to 40--60\% in most trained conditions.

Under untrained lighting, the performance gap remains evident. 
Bi-AQUA maintains high robustness, achieving 100\% in cyan and 80\% in purple and changing illumination, 
whereas the force-ablated policy shows inconsistent behavior (40--100\%). 
These results indicate that force integration significantly improves stability in long-horizon sequential manipulation, 
where perception drift and contact errors may accumulate over multiple stages.

\subsubsection{Peg extraction task: Contact-Rich Peg-out-Hole}
The performance difference becomes more pronounced in the contact-rich peg extraction task. 
Without force input, the policy frequently fails, achieving 0\% success in red, purple, and changing illumination, and at most 80\% even under favorable lighting. 
In contrast, the full Bi-AQUA model achieves 100\% success in all trained lighting modes and maintains high performance under unseen lighting 
(100\% in cyan and changing, 80\% in purple).

The large margin in this task highlights the importance of force-aware action chunking in scenarios with tight geometric tolerance and sustained frictional contact. 
Visual-only control is insufficient when precise alignment and stable interaction are required, particularly under photometric shifts that degrade perception quality.

\subsubsection{Summary}
Overall, the results demonstrate that Bi-AQUA generalizes beyond the original pick-and-place benchmark to both long-horizon sequential manipulation and contact-rich precision tasks in underwater.

\section{Conclusion}
We proposed Bi-AQUA, a bilateral control-based imitation learning framework for underwater robot arms that explicitly models lighting variation within the visuomotor policy. By combining a label-free Lighting Encoder, FiLM-based visual modulation, and a lighting token for transformer conditioning, Bi-AQUA adapts to challenging underwater illumination while preserving the force-sensitive advantages of bilateral control. Experiments on real underwater manipulation tasks showed that Bi-AQUA outperforms a bilateral baseline without lighting modeling and remains robust under static, unseen, and dynamically changing lighting conditions. These results highlight the importance of explicit lighting-aware adaptation for reliable underwater visuomotor learning and suggest a promising path toward practical autonomous underwater manipulation.

\bibliographystyle{IEEEtran}

\end{document}